\documentclass[runningheads]{llncs}
\usepackage{graphicx}

\usepackage{tikz}
\usepackage{comment}
\usepackage{amsmath,amssymb} 
\usepackage{color}

\usepackage{eccv}

\usepackage{multirow}
\usepackage{booktabs}
\usepackage{epstopdf}
\usepackage{cite}

\usepackage{subfig}
\usepackage{array}
\usepackage{mathtools}

\usepackage[pagebackref=true,breaklinks=true,letterpaper=true,colorlinks,bookmarks=false]{hyperref}

\newcolumntype{L}[1]{>{\raggedright\let\newline\\\arraybackslash}m{#1}}
\newcolumntype{C}[1]{>{\centering\let\newline\\\arraybackslash}m{#1}}
\newcolumntype{R}[1]{>{\raggedleft\let\newline\\\arraybackslash}m{#1}}
\newcolumntype{+}{>{\global\let\currentrowstyle\relax}}
\newcolumntype{^}{>{\currentrowstyle}}

\newcommand{\bfp}{\ensuremath{{\mathbf{p}}}}
\newcommand{\bff}{\ensuremath{{\mathbf{f}}}}

\newcommand{\bfW}{\ensuremath{{\mathbf{W}}}}
\newcommand{\bfF}{\ensuremath{{\mathbf{F}}}}
\newcommand{\bfA}{\ensuremath{{\mathbf{A}}}}
\newcommand{\bfP}{\ensuremath{{\mathbf{P}}}}

\newcommand{\calL}{\ensuremath{{\mathcal{L}}}}

\newcommand{\calN}{\ensuremath{{\mathcal{N}}}}
\newcommand{\rom}[1]
    {\MakeUppercase{\romannumeral #1}}

\begin{document}
\pagestyle{headings}
\mainmatter
\def\ECCVSubNumber{2729}  

\title{Interactive Video Object Segmentation Using Global and Local Transfer Modules} 

\titlerunning{Interactive VOS Using Global and Local Transfer Modules}
%
\author{Yuk Heo\inst{1}\orcidID{0000-0002-7425-1254} \and
Yeong Jun Koh\inst{2}\orcidID{0000-0003-1805-2960} \and
Chang-Su Kim\inst{1}\orcidID{0000-0002-4276-1831}}
\authorrunning{Y. Heo, Y. J. Koh, and C.-S. Kim}
%
\institute{School of Electrical Engineering, Korea University, Korea \\
\email{yukheo@mcl.korea.ac.kr changsukim@korea.ac.kr}\\
\and
Department of Computer Science \& Engineering, \\Chungnam National University, Korea\\
\email{yjkoh@cnu.ac.kr}}
\maketitle

\begin{abstract}
An interactive video object segmentation algorithm, which takes scribble annotations on query objects as input, is proposed in this paper. We develop a deep neural network, which consists of the annotation network (A-Net) and the transfer network (T-Net). First, given user scribbles on a frame, A-Net yields a segmentation result based on the encoder-decoder architecture. Second, T-Net transfers the segmentation result bidirectionally to the other frames, by employing the global and local transfer modules. The global transfer module conveys the segmentation information in an annotated frame to a target frame, while the local transfer module propagates the segmentation information in a temporally adjacent frame to the target frame. By applying A-Net and T-Net alternately, a user can obtain desired segmentation results with minimal efforts. We train the entire network in two stages, by emulating user scribbles and employing an auxiliary loss. Experimental results demonstrate that the proposed interactive video object segmentation algorithm outperforms the state-of-the-art conventional algorithms. Codes
and models are available at \href{https://github.com/yuk6heo/IVOS-ATNet}{https://github.com/yuk6heo/IVOS-ATNet}.
\end{abstract}

\section{Introduction}

Video object segmentation (VOS) aims at separating objects of interest from the background in a video sequence. It is an essential technique to facilitate many vision tasks, including action recognition, video retrieval, video summarization, and video editing. Many researches have been carried out to perform VOS, and it can be categorized according to the level of automation. Unsupervised VOS segments out objects with no user annotations, but it may fail to detect objects of interest or separate multiple objects. Semi-supervised VOS extracts target objects, which are manually annotated by a user in the first frame or only a few frames in a video sequence. However, semi-supervised approaches require time-consuming pixel-level annotations (at least 79 seconds per instance as revealed in~\cite{DAVISchallenge2018}) to delineate objects of interest.

\begin{figure}[t]\centering
    \includegraphics[width=\linewidth]{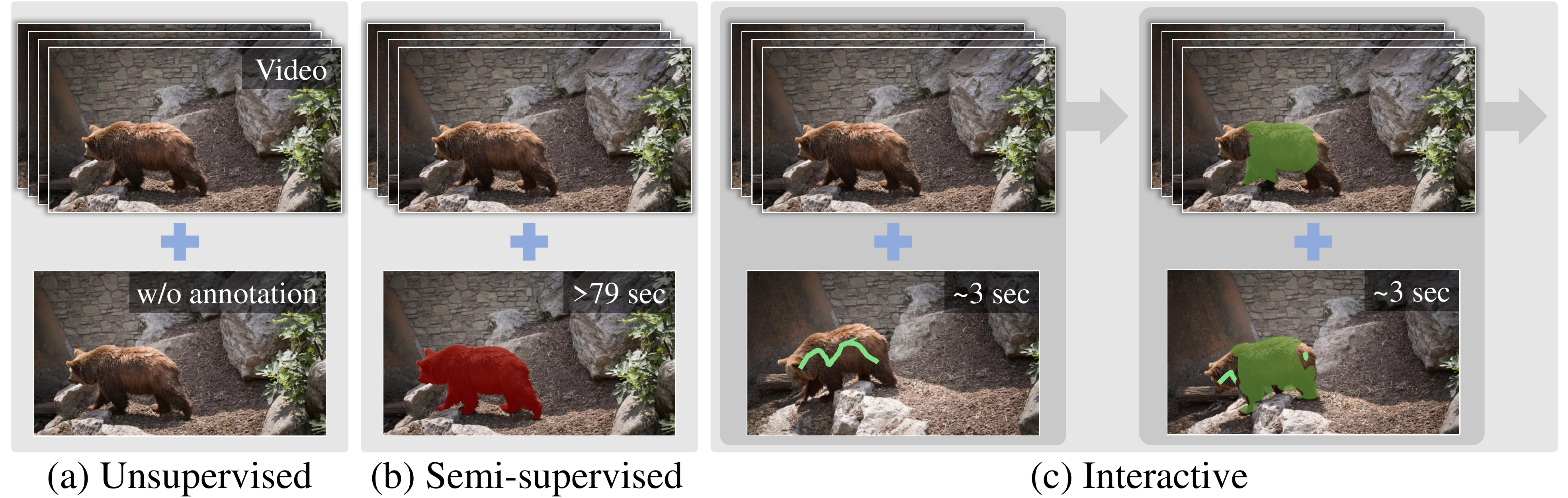}
\caption{Three different levels of supervision in (a) unsupervised VOS, (b) semi-supervised VOS, and (c) interactive VOS. Unsupervised VOS demands no user interaction. Semi-supervised VOS needs pixel-level annotations of an object. Interactive VOS uses quick scribbles and allows interactions with a user repeatedly.}
\label{fig:VOSIntro}
\end{figure}

Therefore, as an alternative approach, we consider interactive VOS, which allows users to interact with segmentation results repeatedly using simple annotations, \eg scribbles, point clicks, or bounding boxes. In this regard, the objective of interactive VOS is to provide reliable segmentation results with minimal user efforts. A work-flow to achieve this objective was presented in the 2018 DAVIS Challenge~\cite{DAVISchallenge2018}. This work-flow employs scribble annotations as supervision, since it takes only about 3 seconds to draw a scribble on an object instance. In this scenario, a user provides scribbles on query objects in a selected frame and the VOS algorithm yields segment tracks for the objects in all frames. We refer to this turn of user-algorithm interaction as a segmentation round. Then, we repeat segmentation rounds to refine the segmentation results until satisfactory results are obtained as illustrated in Fig.~\ref{fig:VOSIntro}(c).

In this paper, we propose a novel approach to achieve interactive VOS using scribble annotations with the work-flow in~\cite{DAVISchallenge2018}. First, we develop the annotation network (A-Net), which produces a segmentation mask for an annotated frame using scribble annotations for query objects. Next, we propose the transfer network (T-Net) to transfer the segmentation result to other target frames subsequently to obtain segment tracks for the query objects. We design the global transfer module and the local transfer module in T-Net to convey segmentation information reliably and accurately. We train A-Net and T-Net in two stages by mimicking scribbles and employing an auxiliary loss. Experimental results verify that the proposed algorithm outperforms the state-of-the-art interactive VOS algorithms on the DAVIS2017~\cite{DAVIS2017}. Also, we perform a user study to demonstrate the effectiveness of the proposed algorithm in real-world applications.

This paper has three main contributions:
\begin{enumerate}
    \item Architecture of A-Net and T-Net with the global and local transfer modules.
    \item Training strategy with the scribble imitation and the auxiliary loss to activate the local transfer module and make it effective in T-Net.
    \item Remarkable performance on the DAVIS dataset in various conditions.
\end{enumerate}

\section{Related Work}

\noindent{\bf Unsupervised VOS:}
Unsupervised VOS is a task to segment out primary objects~\cite{Koh2016pod} in a video without any manual annotations. Before the advance of deep learning, diverse information, including motion, object proposals, and saliency, was employed to solve this problem~\cite{wang2015saliency, papazoglou2013fast, jang2016primary, Koh2017primary, Koh2018}. Recently, many deep learning algorithms with different network architectures have been developed to improve VOS performance using big datasets~\cite{DAVIS2017,Youtube2018}. Tokmakov~\etal~\cite{tokmakov2017learning} presented a fully convolutional model to learn motion patterns from videos. Jain~\etal~\cite{jain2017fusionseg} merged appearance and motion information to perform unsupervised segmentation. Song~\etal~\cite{song2018pyramid} proposed an algorithm using LSTM architecture~\cite{gers1999learning} with atrous convolution layers~\cite{chen2014semantic}. Wang~\etal~\cite{wang2019learning} also adopted LSTM with a visual attention module to simulate human attention.

\noindent{\bf Semi-supervised VOS:}
Semi-supervised VOS extracts query objects using accurately annotated masks at the first frames. Early methods for semi-supervised VOS were developed using hand-crafted features based on random walkers, trajectories, or super-pixels~\cite{brox2010object, jain2014supervoxel, Jang2016BMVC}. Recently, deep neural networks have been adopted for semi-supervised VOS. Some deep learning techniques~\cite{caelles2017one, voigtlaender2017online, maninis2018video} are based on a time-consuming online learning, which fine-tunes a pre-trained network using query object masks at the first frame. Without the fine-tuning, the algorithms in~\cite{Jang2017CVPR, cheng2017segflow, perazzi2017learning, yang2018efficient, Oh2018CVPR} propagate segmentation masks, which are estimated in the previous frame, to the current target frame sequentially for segmenting out query objects. Jang~\etal~\cite{Jang2017CVPR} warped segmentation masks in the previous frame to the target frame and refined the warped masks through convolution trident networks. Yang~\etal~\cite{yang2018efficient} encoded object location information from a previous frame and combined it with visual appearance features to segment out the query object in the target frame. Also, the algorithms in~\cite{chen2018blazingly, hu2018videomatch, voigtlaender2019feelvos, Oh2019ICCV} perform matching between the first frame and a target frame in an embedding space to localize query objects. Chen~\etal~\cite{chen2018blazingly} dichotomized each pixel into either object or background using features from the embedding network. Voigtlaender~\etal~\cite{voigtlaender2019feelvos} trained their embedding network to perform the global and local matching.

\noindent{\bf Interactive image segmentation:}
Interactive image segmentation aims at extracting a target object from the background using user annotations. As annotations, bounding boxes were widely adopted in early methods~\cite{rother2004grabcut, lempitsky2009image, tang2013grabcut, wu2014milcut}. Recently, point-interfaced techniques have been developed~\cite{xu2016deep, ManinisCaelles2018deep, song2018seednet, jang2019interactive}. Maninis~\etal~\cite{ManinisCaelles2018deep} used four extreme points as annotations to inform their network about object boundaries. Jang and Kim~\cite{jang2019interactive} corrected mislabeled pixels through the backpropagating refinement scheme.

\noindent{\bf Interactive VOS:}
Interactive VOS allows users to interact with segmentation results repeatedly using various input types, \eg points, scribbles, and bounding boxes. Users can refine segmentation results until they are satisfied. Some interactive VOS algorithms~\cite{wang2005interactive, price2009livecut, shankar2015video} build graph models using the information in user strokes and segment out target objects via optimization. In~\cite{bai2009video,fan2015jumpcut}, patch matching between target and reference frames is performed to localize query objects. Box interactions can be provided to correct box positions. Benard and Gygli~\cite{benard2017interactive} employed two deep learning networks to achieve interactive VOS. They first obtained object masks from point clicks or scribbles using an interactive image segmentation network and then segmented out the objects using a semi-supervised VOS network. Chen~\etal~\cite{chen2018blazingly} demanded only a small number of point clicks based on pixel-wise metric learning. Oh \etal~\cite{Oh2019CVPR} achieved interactive VOS by following the work-flow in~\cite{DAVISchallenge2018}. They used two segmentation networks to obtain segmentation masks from user scribbles and to propagate the segmentation masks to neighboring frames by exploiting regions of interest. However, their networks may fail to extract query objects outside the regions of interest.

\begin{figure}[t]
\centering
    \includegraphics[width=0.93\linewidth]{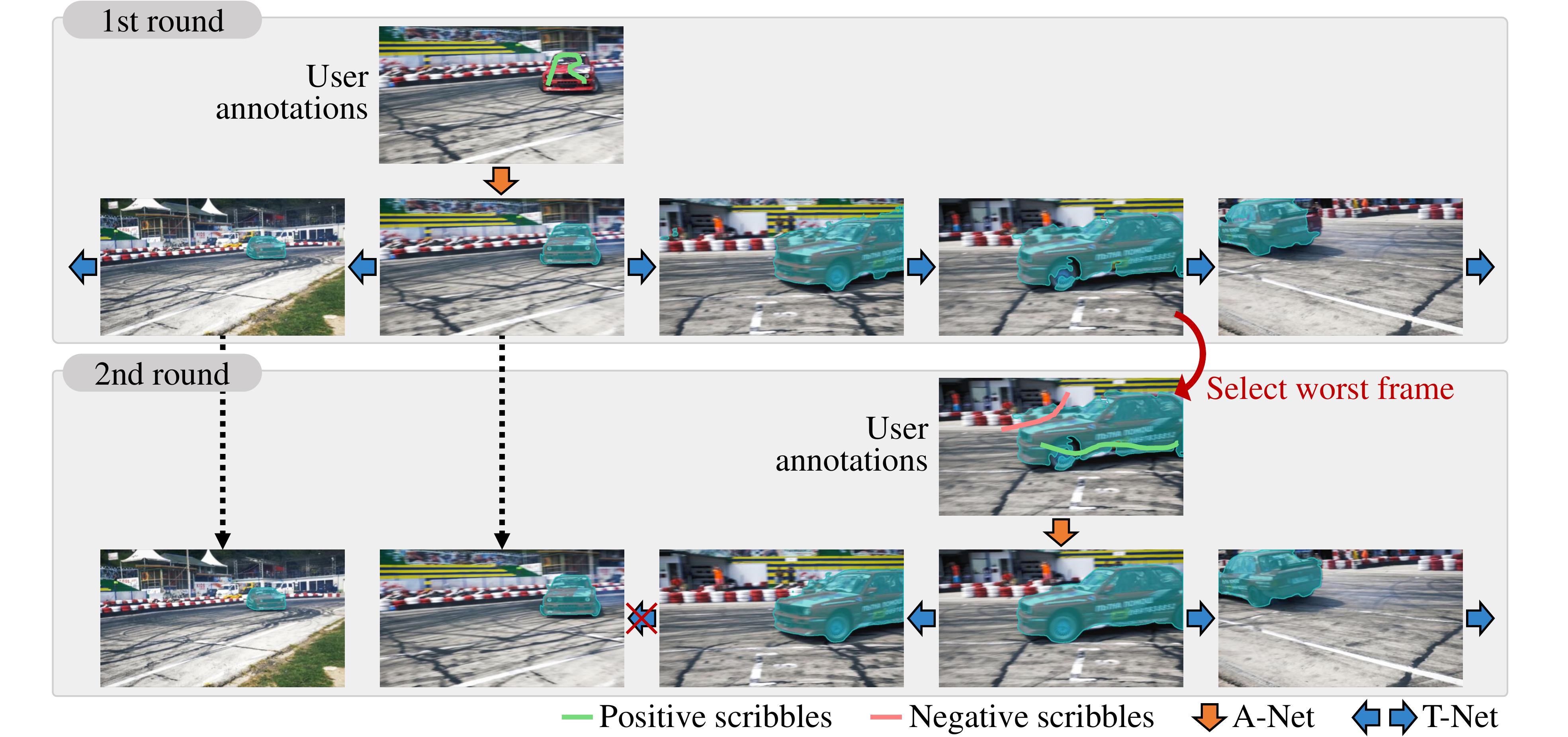}
    \caption{Overview of the proposed interactive VOS algorithm.}
    \label{fig:IVOS_process}
\end{figure}

\section{Proposed Algorithm} \label{sec3:Proposed Algorithm}
We segment out one or more objects in a sequence of video frames through user interactions. To this end, we develop two networks: 1) annotation network (A-Net) and 2) transfer network (T-Net).

Fig.~\ref{fig:IVOS_process} is an overview of the proposed algorithm. In the first segmentation round, a user provides annotations (\ie scribbles) for a query object to A-Net, which then yields a segmentation mask for the annotated frame. Then, T-Net transfers the segmentation mask bi-directionally to both ends of the video to compose a segment track for the object. From the second round, the user selects the poorest segmented frame, and then provides positive and negative scribbles so that A-Net corrects the result. Then, T-Net again propagates the refined segmentation mask to other frames until a previously annotated frame is met. This process is repeated until satisfactory results are obtained.

\subsection{Network architecture} \label{subsec:Network}

\begin{figure}[t]
\centering
    \includegraphics[width=0.95\linewidth]{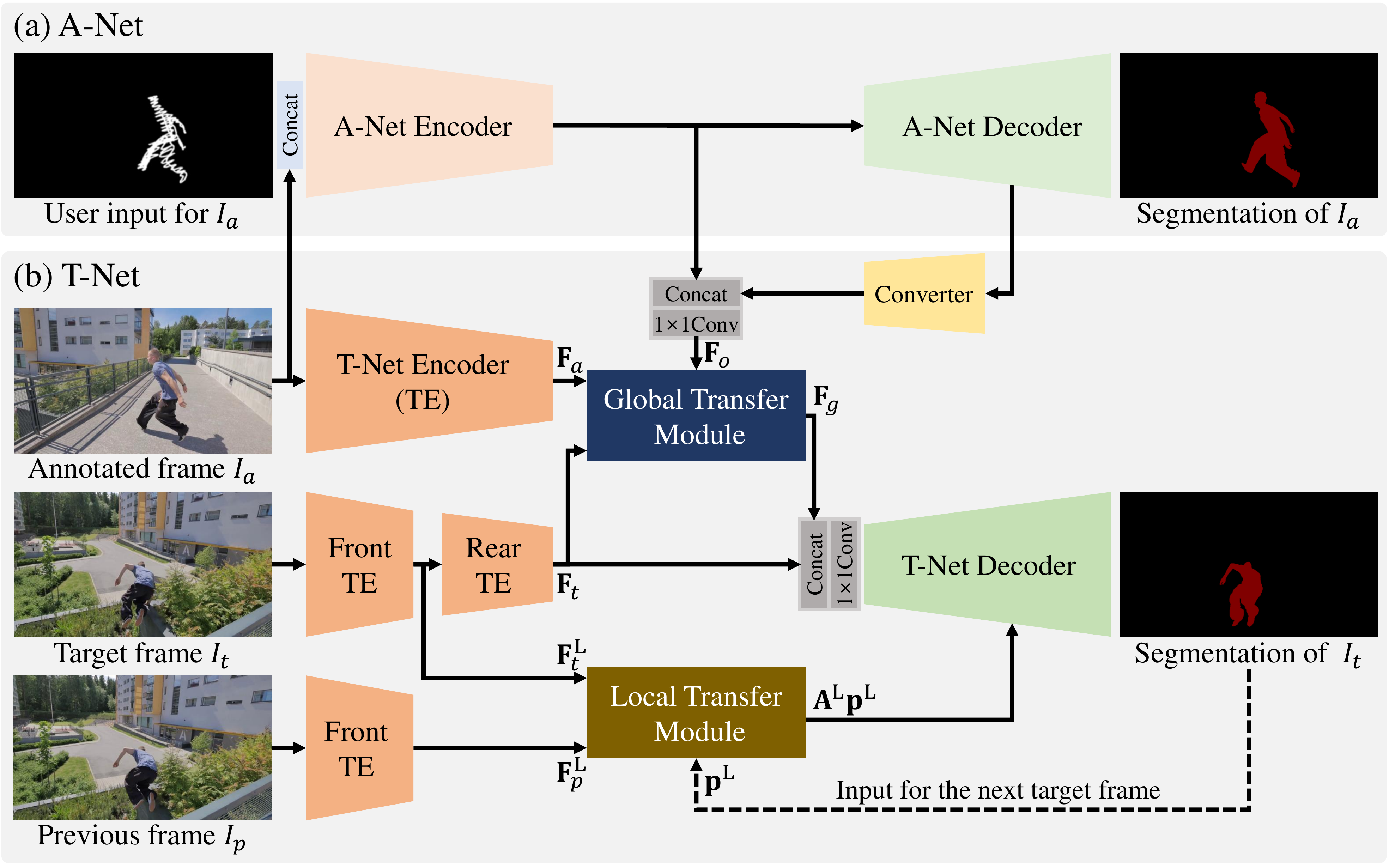}
\caption{Architecture of the proposed networks. A target object in an annotated frame $I_a$ is extracted by A-Net in (a), and the result is sequentially propagated to the other frames, called target frames, by T-Net in (b). In this diagram, skip connections are omitted.}
\label{fig:OverallNetwork}
\end{figure}

Fig.~\ref{fig:OverallNetwork} shows the architecture of the proposed algorithm, which is composed of A-Net and T-Net. First, we segment out query objects in an annotated frame $I_a$ via A-Net. Then, to achieve segmentation in a target frame $I_t$, we develop T-Net, which includes the global and local transfer modules.

\subsubsection{A-Net:} \label{subsubsec:UANet}

Through user interactions, A-Net infers segmentation results in an annotated frame $I_a$. There are two types of interactions according to iteration rounds. In the first round, a user draws scribbles on target objects. In this case, A-Net accepts four-channel input: RGB channels of $I_a$ and one scribble map. In subsequent rounds, the user supplies both positive and negative scribbles after examining the segmentation results in the previous rounds, as illustrated in Fig.~\ref{fig:IVOS_process}. Hence, A-Net takes six channels: RGB channels, segmentation mask map in the previous round, and positive and negative scribble maps. We design A-Net to take 6-channel input, but in the first round, fill in the segmentation mask map with 0.5 and the negative scribble map with 0.

\begin{figure}[h]
\centering
    \includegraphics[width=0.90\linewidth]{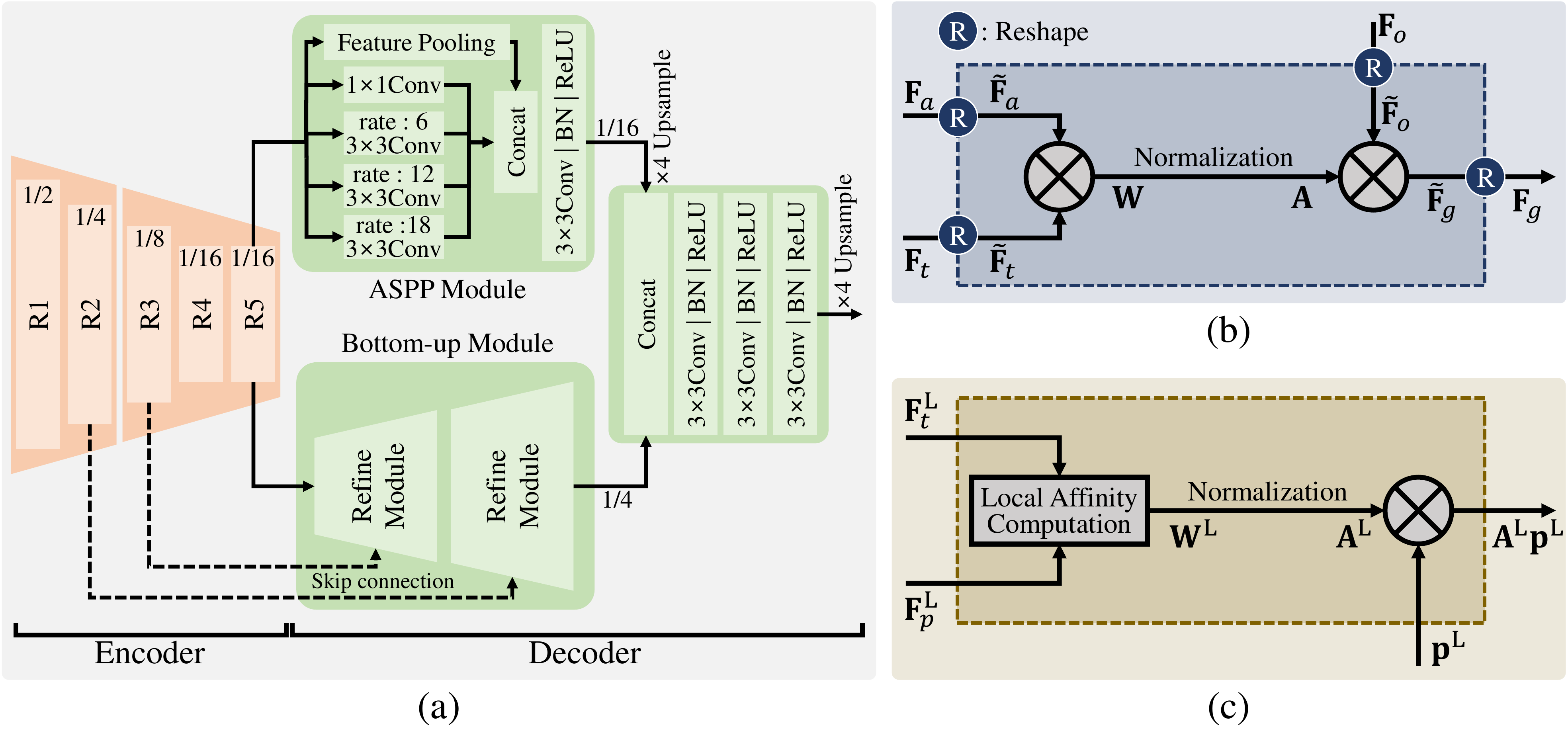}
    \caption{(a) The encoder-decoder architecture, adopted by the proposed A-Net and T-Net. Each fraction is the ratio of the output feature resolution to the input image resolution. (b) Global transfer module. (c) Local transfer module.}
    \label{fig:EncDec}
\end{figure}

A-Net has the encoder-decoder architecture, as specified in Fig.~\ref{fig:EncDec}(a). We adopt SE-ResNet50~\cite{hu2018squeeze} as the encoder to extract features and employ skip connections to consider both low-level and high-level features. We perform dilated convolution and exclude max-pooling in the R5 convolution layer. Then, we use two parallel modules: an ASPP module~\cite{chen2018encoder}, followed by up-sampling with bilinear interpolation, and a bottom-up module. ASPP analyzes multi-scale context features using dilated convolution with varying rates. The bottom-up module consists of two refine modules~\cite{Oh2018CVPR}. The output signals of the ASPP and bottom-up modules are concatenated and then used to predict a probability map of a query object through three sets of convolutional layers, ReLU, and batch normalization. Finally, the estimated probability map is up-sampled to be of the same size as the input image using bilinear interpolation.

\subsubsection{T-Net:} \label{subsubsec:PTNet}
We develop T-Net, which consists of shared encoders, a global transfer module, a local transfer module, and a decoder, as shown in Fig.~\ref{fig:OverallNetwork}(b). The encoders and decoder in T-Net have the same structures as those of A-Net in Fig.~\ref{fig:EncDec}(a). The T-Net decoder yields a probability map for query objects in a target frame $I_t$ using the features from the encoder, the global transfer module, and the local transfer module. Let us describe these two transfer modules.

\subsubsection{Global transfer module:} \label{subsubsec:Transition}
We design the global transfer module to convey the segmentation information of the annotated frame $I_a$ to the target frame $I_t$. Fig.~\ref{fig:EncDec}(b) shows its structure, which adopts the non-local model in~\cite{wang2018non}. It takes two feature volumes $\bfF_t$ and $\bfF_a$ for $I_t$ and $I_a$, respectively. Each volume contains $C$-dimensional feature vectors for $H \times W$ pixels. We then construct an affinity matrix $\bfW$ between $I_t$ and $I_a$, by computing the inner products between all possible pairs of feature vectors in $\bfF_t$ and $\bfF_a$. Specifically, let $\tilde{\bfF}_t\in\mathbb{R}^{HW\times C}$ and $\tilde{\bfF}_a\in\mathbb{R}^{HW\times C}$ denote the feature volumes reshaped into matrices. We perform the matrix multiplication to obtain the affinity matrix
\begin{equation}
\bfW=\tilde{\bfF}_t \times \tilde{\bfF}_a^T.
\end{equation}
Its element $\bfW(i,j)$ represents the affinity of the $i$th pixel in $\tilde{\bfF}_t$ to the $j$th pixel in $\tilde{\bfF}_a$. Then, we obtain the transition matrix $\bfA$ by applying the softmax normalization to each column in $\bfW$.

The transition matrix $\bfA$ contains matching probabilities from pixels in $I_a$ to those in $I_t$. Therefore, it can transfer query object probabilities in $I_a$ to $I_t$. To approximate these probabilities in $I_a$, we extract a mid-layer feature from the A-Net decoder, down-sample it using the converter, which includes two sets of SE-Resblock~\cite{hu2018squeeze} and max-pooling layer. Then, its channels are halved by 1$\times$1 convolutions after it is concatenated to the output of the A-Net encoder, as shown in Fig.~\ref{fig:OverallNetwork}. The concatenated feature $\bfF_o$ is fed into the global transfer module, as shown in Fig.~\ref{fig:EncDec}(b). Then, it is reshaped into $\tilde{\bfF}_o$, which represents the query object feature distribution in $I_a$. Finally, the global transfer module produces the transferred distribution
\begin{equation}
\tilde{\bfF}_g = \bfA \tilde{\bfF}_o,
\end{equation}
which can be regarded as an inter-image estimate of the query object feature distribution in $I_t$. Then the distribution is reshaped into $\bfF_g\in\mathbb{R}^{H\times W\times C}$ to be input to the T-Net decoder.

From the second round, there are $N$ annotated frames, where $N$ is the ordinal index for the round. To obtain reliable segmentation results, we use all information in the $N$ annotated frames. Specifically, we compute the transition matrix $\bfA^{(i)}$ from the $i$th annotated frame to $I_t$ and the query object distribution $\bfF_o^{(i)}$ in the $i$th annotated frame. Then, we obtain the average of the multiple inter-image estimates of the query object distribution in $I_t$ by
\begin{equation}
\tilde{\bfF}_g = \frac{1}{N} \sum_{i=1}^{N} \bfA^{(i)} \tilde{\bfF}_o^{(i)}.
\end{equation}

\subsubsection{Local transfer module:} \label{subsubsec:RFCmodule}
The segmentation information in an annotated frame is propagated bidirectionally throughout the sequence. Thus, during the propagation, when a target frame $I_t$ is to be segmented, there is the previous frame $I_p$ that is already segmented. We design the local transfer module to convey the segmentation information in $I_p$ to $I_t$.

The local transfer module is similar to the global one, but it performs matching locally since $I_t$ and $I_p$ are temporally adjacent. In other words, object motions between $I_t$ and $I_p$, which tend to be smaller than those between $I_t$ and $I_a$, are estimated locally. Furthermore, since $I_t$ and $I_p$ are more highly correlated, motions between them can be estimated more accurately. Therefore, the local module uses higher-resolution features than the global one does. Specifically, the local module takes features from the R2 convolution layer in the encoder in Fig.~\ref{fig:EncDec}(a), instead of the R5 layer. $\bfF_t^{\rm L}$ and $\bfF_p^{\rm L}$, which denote these feature volumes from $I_t$ and $I_p$, are provided to the local transfer module, as shown in Fig.~\ref{fig:EncDec}(c). Then, we compute the local affinity matrix $\bfW^{\rm L}$, whose $(i,j)$th element indicates the similarity between the $i$th pixel $I_t$ and the $j$th pixel in $I_p$. Specifically, $\bfW^{\rm L}(i,j)$ is defined as
 \begin{align}
    {\bfW^{\rm L}(i,j)}
        =\left\{
            \begin{array}{ll}
                \bff_{t,i}^T \bff_{p,j} \quad \quad  & j\in \calN_i,\\
                0                       \quad \quad   & \mbox{otherwise},
            \end{array}
        \right.
    \label{eq:matchcost}
\end{align}
where $\bff_{t,i}$ and $\bff_{p,j}$ are the feature vectors for the $i$th pixel in $\bfF_t^{\rm L}$ and the $j$th pixel in $\bfF_p^{\rm L}$, respectively. Also, the local region $\calN_i$ is the set of pixels, which are sampled from $(2d+1)\times (2d+1)$ pixels around pixel $i$ with stride 2 to reduce the computations. In this work, $d$ is set to 4. Then, the affinity is computed for those pixels in the local region only, and set to be zeros for the other pixels.

As in the global module, $\bfW^{\rm L}$ is normalized column-by-column to the transition matrix $\bfA^{\rm L}$. Also, a segmentation mask map $\bfP_{p}$ in the previous frame $I_p$ is down-sampled and vectorized to obtain a probability vector $\bfp^{\rm L}$. Then, we obtain $\bfA^{\rm L}\bfp^{\rm L}$, which is another estimate of the query object distribution in $I_t$. It has a higher resolution than the estimate in the global module, and thus is added to the corresponding mid-layer in the T-Net decoder, as shown in Fig.~\ref{fig:OverallNetwork}(b).

Computing global and local similarities in the proposed global and local transfer modules is conceptually similar to~\cite{voigtlaender2019feelvos}, but their usage is significantly different. Although~\cite{voigtlaender2019feelvos} also computes global and local distances, it transforms those distances into a single channel by taking the minimum distance at each position. Thus, it loses a substantial amount of distance information. In contrast, the proposed algorithm computes global and local affinity matrices and uses them to transfer object probabilities from annotated and previous frames to a target frame. In Section~\ref{subsec:AblationStudy}, we verify that the proposed global and local modules are more effective than the best matching approach in~\cite{voigtlaender2019feelvos}.

\subsection{Training phase} \label{subsec:Training}

We train the proposed interactive VOS networks in two stages, since T-Net should use A-Net output; we first train A-Net and then train T-Net using the trained A-Net.

\subsubsection{A-Net training:} To train A-Net, we use the image segmentation dataset in~\cite{Hariharan2011} and the video segmentation datasets in~\cite{DAVIS2017, Youtube2018}. Only a small percentage of videos in the DAVIS2017 dataset~\cite{DAVIS2017} provide user scribble data. Hence, we emulate user scribbles via two schemes: 1) point generation and 2) scribble generation in~\cite{DAVISchallenge2018}.

In the first round, A-Net yields a segmentation mask for a query object using positive scribbles only. We perform the point generation to imitate those positive scribbles. We produce a point map by sampling points from the ground-truth mask for the query object. Specifically, we pick one point randomly for every $100 \sim 3000$ object pixels. We vary the sampling rate to reflect that users provide scribbles with different densities. Then, we use the generated point map as the positive scribble map.

In each subsequent round, A-Net should refine the segmentation mask in the previous round using both positive and negative scribbles. To mimic an inaccurate segmentation mask, we deform the ground-truth mask using various affine transformations. Then, we extract positive and negative scribbles using the scribble generation scheme in~\cite{DAVISchallenge2018}, by comparing the deformed mask with the ground-truth. Then, $I_a$, the deformed mask, and the generated positive and negative scribble maps are fed into A-Net for training.

We adopt the pixel-wise class-balanced cross-entropy loss~\cite{xie2015holistically} between A-Net output and the ground-truth. We adopt the Adam optimizer to minimize this loss for 60 epochs with a learning rate of $1.0\times10^{-5}$. We decrease the learning rate by a factor of 0.2 every 20 epochs. In each epoch, the training is iterated for 7,000 mini-batches, each of which includes 6 pairs of image and ground-truth. For data augmentation, we apply random affine transforms to the pairs.

\subsubsection{T-Net training:}
For each video, we randomly pick one frame as an annotated frame, and then select seven consecutive frames, adjacent to the annotated frame, in either the forward or backward direction. Among those seven frames, we randomly choose four frames to form a mini-sequence. Thus, there are five frames in a mini-sequence: one annotated frame and four target frames. For each target frame
in the mini-sequence, we train T-Net using the features from the trained A-Net, which takes the annotated frame as input.

We compare an estimated segmentation mask with the ground-truth to train T-Net, by employing the loss function
\begin{equation}
    \calL = \calL_{\rm c} + \lambda_{\rm}\calL_{\rm aux}
    \label{eq:TNetloss}
\end{equation}
where $\calL_{\rm c}$ is the pixel-wise class-balanced cross-entropy loss between the T-Net output and the ground-truth. The auxiliary loss $\calL_{\rm aux}$ is the pixel-wise mean square loss between the transferred probability map, which is the output of the local transfer module, and the down-sampled ground-truth. The auxiliary loss $\calL_{\rm aux}$ enforces the front encoders of T-Net in Fig.~\ref{fig:OverallNetwork} to generate appropriate features for transferring the previous segmentation mask successfully. Also, $\lambda_{\rm}$ is a balancing hyper-parameter, which is set to 0.1. We also employ the Adam optimizer to minimize the loss function for 40 epochs with a learning rate of $1.0\times10^{-5}$, which is decreased by a factor of 0.2 every 20 epochs. The training is iterated 6,000 mini-batches, each of which contains 8 mini-sequences.

\subsection{Inference phase} \label{subsec:Inference}
Suppose that there are multiple target objects. In the first round, for each target object in an annotated frame, A-Net accepts the user scribbles on the object and produces a probability map for the object. To obtain multiple object segmentation results, after zeroing probabilities lower than 0.8, each pixel is assigned to the target object class, corresponding to the highest probability. Then, T-Net transfers the multiple segmentation masks in the annotated frame bi-directionally to both ends of the sequence. T-Net also compares the multiple probability maps and determines the target object class of each pixel, as done in A-Net. From the second round, the user selects the frame with the poorest segmentation results and then provides additional positive and negative scribbles. The scribbles are then fed into A-Net to refine the segmentation results. Then, we transfer segmentation results bidirectionally with T-Net. In each direction, the transmission is carried out until another annotated frame is found.

During the transfer, we superpose the result of segmentation mask ${\bfP}_t^{r}$ for frame $I_t$ in the current round $r$ with that $\bfP_{t}^{r-1}$ in the previous round. Specifically, the updated result $\tilde{\bfP}_t^r$ in round $r$ is given by

\begin{equation}
    \tilde{\bfP}_t^r=\frac{1}{2}(1+\frac{t-t_{b}}{t_{r}-t_{b}})\bfP_{t}^r + \frac{t_{r}-t}{2(t_{r}-t_{b})}{\bfP}_t^{r-1}
\end{equation}
where $t_{r}$ is the annotated frame in round $r$ and $t_b$ is one of the previously annotated frames, which is the closest to $t$ in the direction of the transfer. By employing this superposition scheme, we can reduce drifts due to a long temporal distance between annotated and target frames.

\begin{figure}[t]
\centering
    \subfloat{\includegraphics[width=0.40\linewidth]{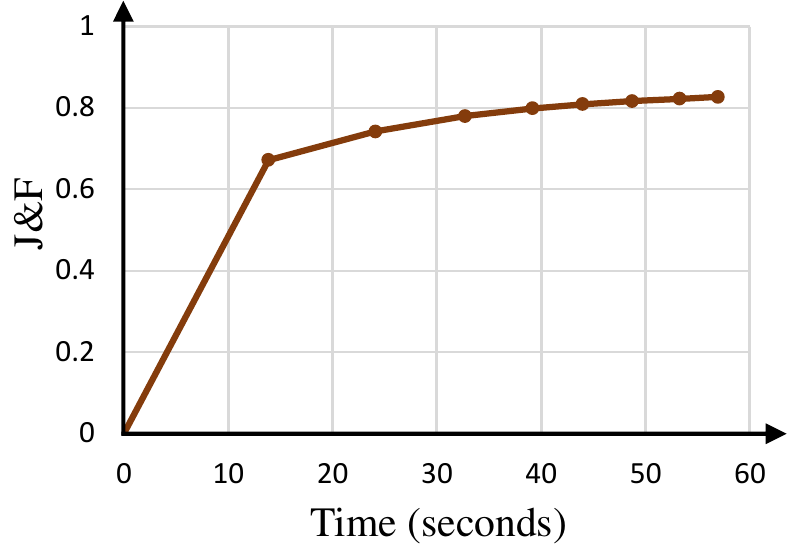}}
    \subfloat{\includegraphics[width=0.40\linewidth]{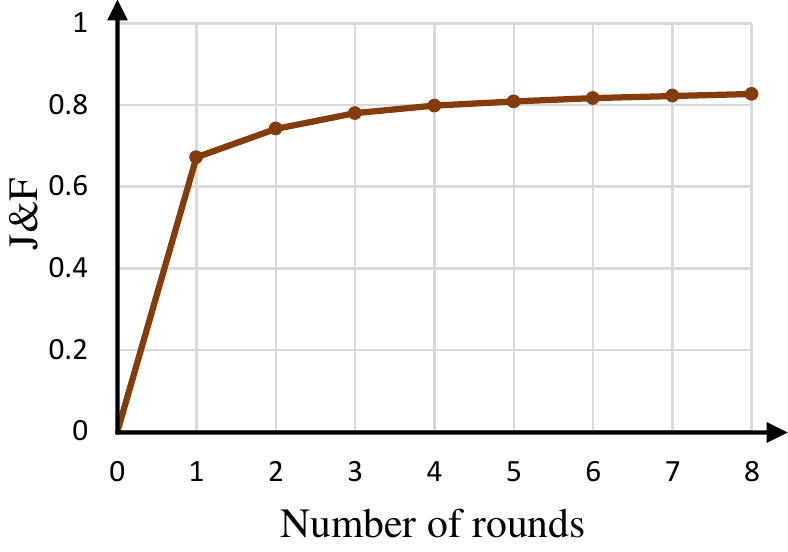}}
\caption{J\&F performances of the proposed algorithm on the validation set in DAVIS2017 according to the time and the number of rounds.}
\label{fig:graph}
\end{figure}

\begin{table}[t]\footnotesize\centering
\caption{Comparison of the proposed algorithm with the conventional algorithms on the DAVIS2017 validation set. The best results are boldfaced.}

\begin{tabular}[t]{+L{3.0cm}^C{1.5cm}^C{1.5cm}^C{1.5cm}^C{1.5cm}} \toprule
                             & AUC-J    & J@60s     & AUC-J\&F& J\&F@60s\\ \midrule
Najafi~\etal~\cite{DAVIS2018IVOS2nd}    & 0.702      & 0.548      & $-$          & $-$ \\
Heo~\etal~\cite{DAVIS2019IVOS2nd}       & 0.704      & 0.725      & 0.734        & 0.752\\
Ren~\etal~\cite{DAVIS2019IVOS4th}       & $-$        & $-$        & 0.766        & 0.780\\
Oh~\etal~\cite{Oh2019CVPR}              & 0.691      & 0.734      & $-$          & $-$\\
Proposed                                & \bf{0.771} & \bf{0.790} & \bf{0.809}  & \bf{0.827}\\ \bottomrule
\end{tabular}
\label{tb:ComparisonAuto}
\end{table}

\section{Experimental Results} \label{sec4:experiments}
We first compare the proposed interactive VOS algorithm with conventional algorithms. Second, we conduct a user study to assess the proposed algorithm in real-world applications. Finally, we do various ablation studies to analyze the proposed algorithm.

\subsection{Comparative assessment} \label{subsec:Experi}
In this test, we follow the interactive VOS work-flow in~\cite{DAVISchallenge2018}. The work-flow first provides a manually generated scribble for each target object in the first round, and then automatically generates additional positive and negative scribbles to refine the worst frames in up to 8 subsequent rounds. There are three different scribbles provided in the first round. In other words, three experiments are performed for each video sequence. The region similarity (J) and contour accuracy (F) metrics are employed to assess VOS algorithms. For the evaluation of interactive VOS, we measure the area under the curve for J score (AUC-J) and for joint J and F scores (AUC-J\&F) to observe the overall performance according over the 8 segmentation rounds. Also, we measure the J score at 60 seconds (J@60s), and the joint J and F score at 60 seconds (J\&F@60s) to evaluate how much performance is achieved within the restricted time.

\begin{figure}[t]
\centering
    \includegraphics[width=\linewidth]{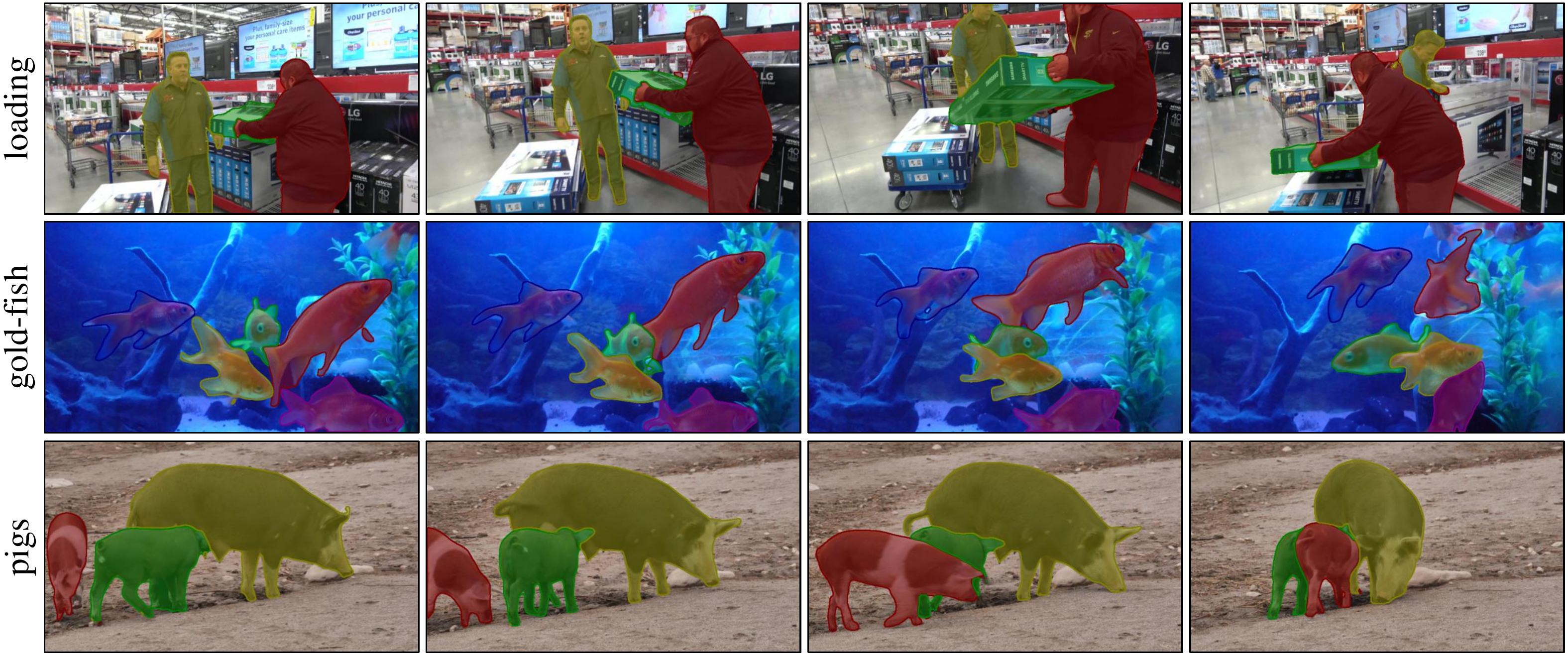}
\caption{Results of the proposed interactive VOS algorithm after 8 rounds.}
\label{fig:Automatic}
\end{figure}

\begin{figure}[t]
\centering
   \includegraphics[width=70mm]{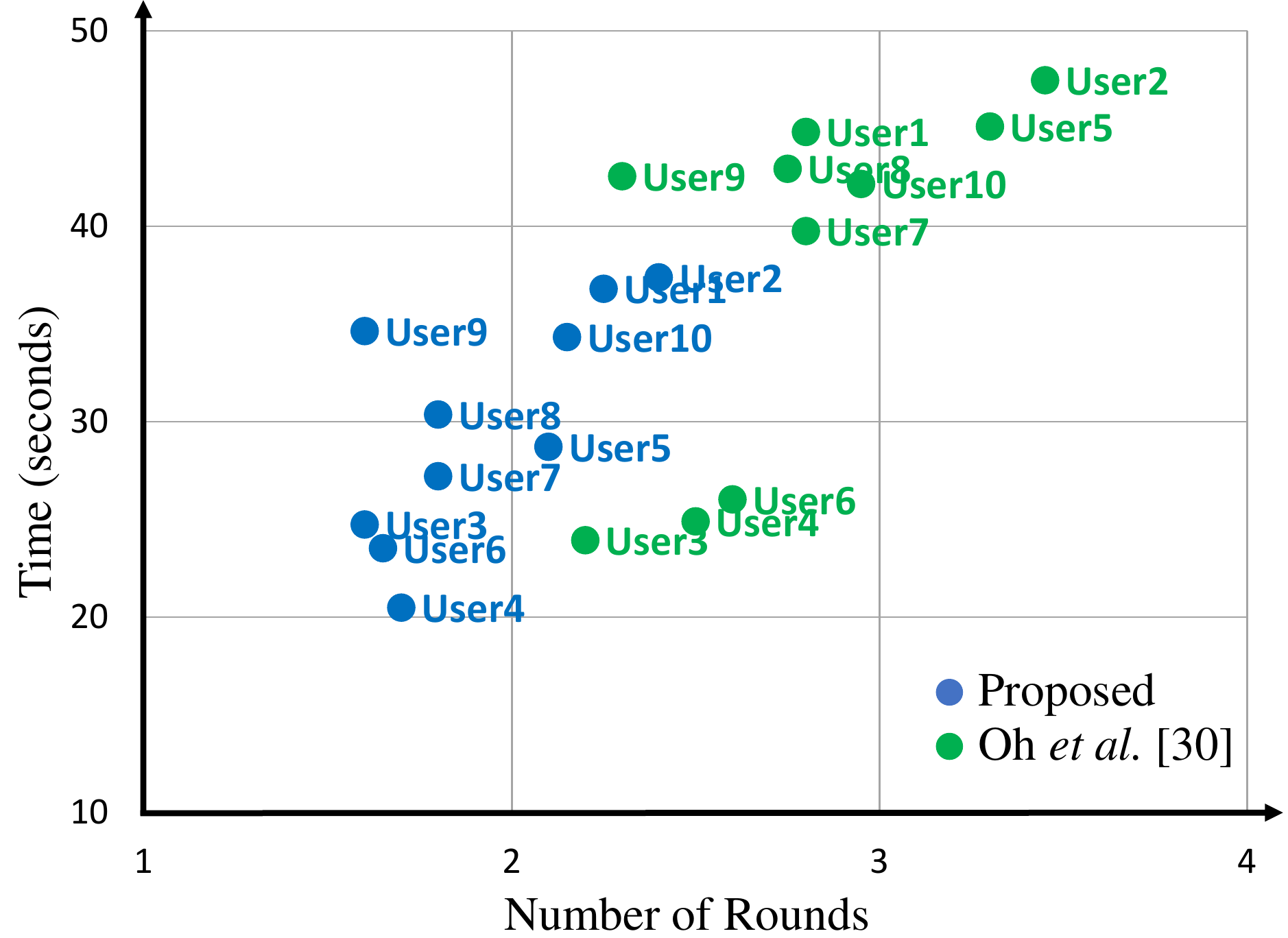}
\caption{Comparison of the average times and average round numbers.}
\label{fig:GraphUser}
\end{figure}

Fig.~\ref{fig:graph} shows the J\&F performances of the proposed algorithm on the validation set in DAVIS2017~\cite{DAVIS2017} according to the time and the number of rounds. The performances increase quickly and saturate at around 40s or in the third round. Also, we observe that the 8-round experiment is completed within 60 seconds. Table~\ref{tb:ComparisonAuto} compares the proposed algorithm with recent state-of-the-art algorithms~\cite{DAVIS2018IVOS2nd,DAVIS2019IVOS2nd,DAVIS2019IVOS4th,Oh2019CVPR}. The scores of the conventional algorithms are from the respective papers. The proposed algorithm outperforms the conventional algorithms by significant margins in all metrics. Fig.~\ref{fig:Automatic} presents examples of segmentation results of the proposed algorithm after 8 rounds. We see that multiple primary objects are segmented out faithfully.

\subsection{User study}

\begin{table}[t]\footnotesize\centering
\caption{Summary of the user study results.}
\begin{tabular}[t]{+L{2.3cm}^C{1.5cm}^C{1.5cm}^C{1.5cm}^C{1.5cm}}
\toprule
                    & SPV    & RPV      & J Mean    & F Mean  \\
\midrule
Oh~\etal~\cite{Oh2019CVPR}      & 37.9   & 2.77     & 0.823                 & 0.817  \\
Proposed                        & \bf{29.8}   & \bf{1.90}   & \bf{0.832}    & \bf{0.822}  \\
\bottomrule
\end{tabular}
\label{tb:ComparisonManual}
\end{table}

\begin{figure}[t]
\centering
    \includegraphics[width=\linewidth]{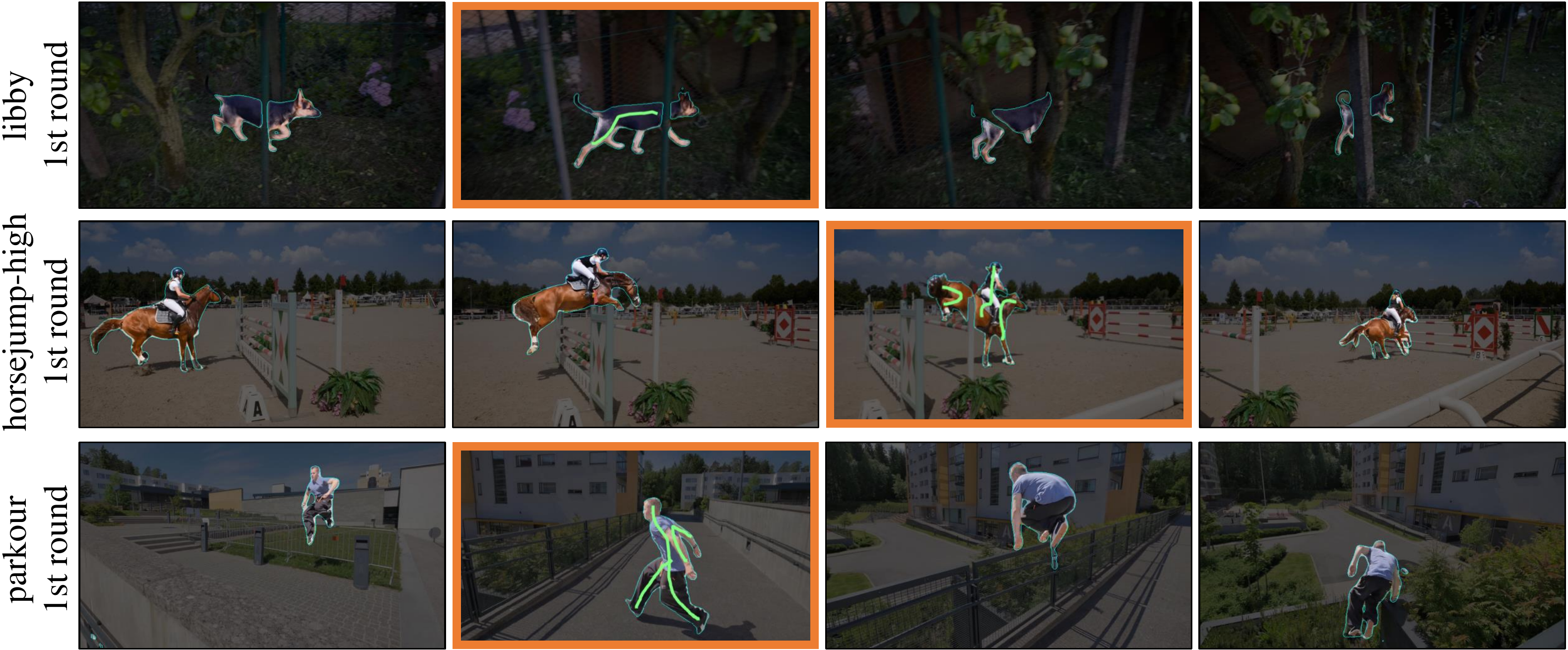}
\caption{Examples of scribbles and segmentation results during the user study. Positive and negative scribbles are depicted in green and red, respectively.}
\label{fig:Manual}
\end{figure}

We conduct a user study, by recruiting 10 off-line volunteers and asking them to provide scribbles repeatedly until they are satisfied. We measure the average time in seconds per video (SPV), including the interaction time to provide scribbles and the running time of the algorithm, and the average round number in rounds per video (RPV) until the completion. Also, we report the J and F means of all sequences when the interactive process is completed.

We perform the user study for the proposed algorithm and the state-of-the-art interactive VOS algorithm~\cite{Oh2019CVPR}. For this comparison, we use the validation set (20 sequences) in DAVIS2016~\cite{perazzi2016DAVIS}, in which each video contains only a single query object. This is because the provided source code of~\cite{Oh2019CVPR} works on a single-object case only. Fig.~\ref{fig:GraphUser} plots the average time and the average round number for each user. We observe that all users, except user 3, spend less time and conduct fewer rounds using the proposed algorithm. Table~\ref{tb:ComparisonManual} summarizes the user study results. The proposed algorithm is faster than~\cite{Oh2019CVPR} in terms of both SPV and RPV. It is worth pointing out that the proposed algorithm yields better segmentation results within shorter times.

Fig.~\ref{fig:Manual} shows examples of segmentation results in the user study. For the ``Libby,'' ``Horsejump-High,'' and ``Parkour'' sequences, the proposed algorithm deals with occlusions and scale changes of query objects effectively, and completes the segmentation in just a single round. Please see the supplemental video to see how the evaluation works.

\begin{table}[t]\footnotesize\centering
\caption{Ablation study on the local transfer module (J scores on the validation set in DAVIS2017).}
\begin{tabular}[t]{+C{2.2cm}+C{1.5cm}+C{1.5cm}+C{1.2cm}+C{0.001cm}+C{0.95cm}^C{0.95cm}^C{0.95cm}^C{0.95cm}^C{0.95cm}}
\toprule
& & & && \multicolumn{5}{c}{Round}\\
\cmidrule(lr){6-10}
Method & Front TE & Rear TE &$\lambda$ && 1st   & 2nd   & 3rd   & 4th & 5th\\
\midrule
\rom{1} & \multicolumn{3}{c}{w/o local transfer module} && 0.629 & 0.704 & 0.741 & 0.759 & 0.771 \\
\rom{2} & & \checkmark & 0.1 && 0.653 & 0.708 & 0.738 & 0.751 & 0.760 \\
\rom{3} & \checkmark & & 0.0 && 0.645 & 0.706 & 0.735 & 0.750 & 0.761 \\
\rom{4} & \checkmark & & 0.5 && 0.658 & 0.721 & 0.748 & 0.758 & 0.772 \\
\rom{5} & \checkmark & & 1.0 && 0.654 & 0.715 & 0.742 & 0.755 & 0.762 \\
\rom{6} (Proposed) & \checkmark & & 0.1 && \bf{0.676} & \bf{0.732} & \bf{0.762} & \bf{0.772} & \bf{0.783}\\
\bottomrule
\end{tabular}
\label{tb:Ablation}
\end{table}

\subsection{Ablation studies}
\label{subsec:AblationStudy}
We analyze the efficacy of the proposed global and local transfer modules through two ablation studies.

First, we verify that the structure and the training method of the local transfer module are effective. In Table~\ref{tb:Ablation}, we report the J scores on the validation set in DAVIS2017, by varying the configurations of the local transfer module. In method \rom{1}, we assess the proposed algorithm without the local transfer module. Note that the J scores in early rounds degrade severely. The local model is hence essential for providing satisfactory results to users quickly in only a few rounds. Method \rom{2} uses the features of rear TE, instead of those of front TE to compute the affinity matrix of the local transfer module. The features of the front TE are more effective than those of rear TE because of their higher spatial resolution. In method \rom{3}, without the auxiliary loss ${\cal L}_{\rm aux}$ (\ie $\lambda=0$ in \eqref{eq:TNetloss}), the local transfer module becomes ineffective and the performances degrade significantly. Methods \rom{4}, \rom{5}, and \rom{6} vary the parameter $\lambda$. We see that $\lambda=0.1$ performs the best by balancing the two losses in \eqref{eq:TNetloss}.

\begin{table}[t]\footnotesize\centering
\caption{Ablation study to validate the proposed probability transfer approach.}
\begin{tabular}[t]{+L{6.5cm}^C{1.15cm}^C{1.15cm}^C{1.5cm}^C{1.4cm}}
\toprule
                                               & AUC-J      & J@60s      & AUC-J\&F    & J\&F@60s\\
\midrule
Matching approach~\cite{voigtlaender2019feelvos} {\scriptsize (predictions of A-Net)}       & 0.636      & 0.653      & 0.654       & 0.670 \\
Matching approach~\cite{voigtlaender2019feelvos} {\scriptsize (scribble annotations)}       & 0.661      & 0.676      & 0.674           & 0.690\\
Proposed probability transfer approach                                                     & \bf{0.771} & \bf{0.790} & \bf{0.809}  & \bf{0.827}\\
\bottomrule
\end{tabular}
\label{tb:AblationMatch}
\end{table}

Next, we verify that the proposed global and local transfer modules are more effective for interactive VOS than the global and local matching in~\cite{voigtlaender2019feelvos}. Note that \cite{voigtlaender2019feelvos} is a semi-supervised VOS algorithm, which estimates matching maps between a target frame and the target object region. We plug its matching modules into the proposed interactive system. More specifically, we compute a global similarity map between a target frame and the target object region in an annotated frame to perform the global matching in~\cite{voigtlaender2019feelvos}. We determine the target object region in two ways: 1) the region predicted by A-Net or 2) the set of scribble-annotated pixels. We then transform the similarity map into a single channel by taking the maximum similarity at each position. Then, we replace $\bfF_g$, which is the output of the proposed global transfer module, with the single-channel similarity. For the local matching, we obtain a local similarity map between the target frame and the segmentation region in the previous frame to compose another single-channel similarity. We then feed the local matching result, instead of $\bfA^{\rm L}\bfp^{\rm L}$, to the T-Net decoder. We train these modified networks using the same training set as the proposed networks. The implementation details of the modified networks can be found in the supplemental document. Table~\ref{tb:AblationMatch} compares the performances of the proposed transfer modules with those of the matching modules in~\cite{voigtlaender2019feelvos} on the validation set in DAVIS2017. We observe that the proposed probability transfer approach outperforms the best matching approach~\cite{voigtlaender2019feelvos} significantly.

\section{Conclusions} \label{sec5:conclusion}

We proposed a novel interactive VOS algorithm using A-Net and T-Net. Based on the encoder-decoder architecture, A-Net processes user scribbles on an annotated frame to generate a segmentation result. Then, using the global and local transfer modules, T-Net conveys the segmentation information to the other frames in the video sequence. These two modules are complementary to each other. The global module transfers the information from an annotated frame to a target frame reliably, while the local module conveys the information between adjacent frames accurately. In the training process, we introduced the point-generation method to compensate for the lack of scribble-annotated data. Moreover, we incorporated the auxiliary loss to activate the local transfer module and make it effective in T-Net. By employing A-Net and T-Net repeatedly, a user can obtain satisfactory segmentation results. Experimental results showed that the proposed algorithm performs better than the state-of-the-art algorithms, while requiring fewer interaction rounds.

\section*{Acknowledgements}

This work was supported in part by `The Cross-Ministry Giga KOREA Project' grant funded by the Korea government (MSIT) (No.GK20P0200, Development of 4D reconstruction and dynamic deformable action model based hyper-realistic service technology), in part by Institute of Information \& communications Technology Planning \& evaluation (IITP) grant funded by the Korea government (MSIT) (No.2020-0-01441, Artificial Intelligence Convergence Research Center (Chungnam National University)) and in part by the National Research Foundation of Korea (NRF) through the Korea Government (MSIP) under Grant NRF-2018R1A2B3003896.

\bibliographystyle{splncs04}
\bibliography{2729}
\end{document}